%% file: example_paper.tex
\documentclass[11pt]{article}

\usepackage{microtype}
\usepackage{graphicx}
\usepackage{subcaption}
\usepackage{booktabs} 
\usepackage{times}
\usepackage{latexsym}
\usepackage[T1]{fontenc}
\usepackage[utf8]{inputenc}
\usepackage{inconsolata}

\usepackage{hyperref}

\usepackage[preprint]{acl}

\usepackage{amsmath}
\usepackage{amssymb}
\usepackage{mathtools}
\usepackage{amsthm}
\usepackage{dsfont}
\usepackage{multirow}
\usepackage{arydshln}
\usepackage{bookmark}
\usepackage{algorithmic}
\usepackage[ruled,vlined]{algorithm2e}
\SetAlFnt{\small}
\SetAlCapFnt{\small}
\SetAlCapNameFnt{\small}

\usepackage[table]{xcolor}  
\definecolor{customTeal}{RGB}{0, 128, 128} 
\definecolor{emphasisColor}{RGB}{255, 0, 0} 
\definecolor{oursBlue}{RGB}{51,202,246}

\definecolor{blue1}{HTML}{508AB2}
\definecolor{green2}{HTML}{BFF6BA}
\definecolor{RoyalBlue}{RGB}{65, 105, 225}

\usepackage[capitalize,noabbrev]{cleveref}

\theoremstyle{plain}

\theoremstyle{definition}

\theoremstyle{remark}

\usepackage[textsize=tiny]{todonotes}

\usepackage{xcolor}      
\usepackage{amsmath}     
\usepackage{amssymb}     
\usepackage{tcolorbox}   
\tcbuselibrary{theorems,breakable} 

\usepackage{xcolor}      
\usepackage{amsmath}     
\usepackage{amssymb}     
\usepackage{tcolorbox}   
\usepackage[T1]{fontenc}
\tcbuselibrary{theorems,breakable} 

\usepackage{wrapfig}

\definecolor{SeaGreen}{RGB}{46,139,87}
\definecolor{CornflowerBlue}{RGB}{100,149,237}
\definecolor{RoyalPurple}{RGB}{120,81,169}
\definecolor{Aquamarine}{RGB}{127,255,212}

\newtcbtheorem[]{prompt}{Prompt}%
{colback=SeaGreen!10!CornflowerBlue!10,
 colframe=RoyalPurple!55!Aquamarine!100!,
 fonttitle=\bfseries,
 left=.02in, right=.02in, bottom=.02in, top=.02in,
 before upper={\linespread{1.5}\selectfont}}{prompt}

\title{Adaptive Test-Time Compute Allocation for Block Diffusion Language Models in Complex Reasoning}

\author{
\textbf{Yi Lu}$^{1,3}$\thanks{Equal contribution.},
\quad
\textbf{Deyang Kong}$^{2,3}$\footnotemark[1],
\quad
\textbf{Jianing Wang}$^{3}$,
\quad
\textbf{Linsen Guo}$^{3}$,
\quad
\textbf{Xue Wang}$^{3}$
\quad
\textbf{Qi Guo}$^{2,3}$,
\\
\textbf{Tao Gui}$^{1}$\thanks{Corresponding authors.},
\quad
\textbf{Xuanjing Huang}$^{1}$,
\quad
\textbf{Wei Ye}$^{2}$\footnotemark[2],
\quad
\textbf{Shikun Zhang}$^{2}$,
\quad
\textbf{Wei Wang}$^{3}$\footnotemark[2]
\\
$^1$Fudan University
\quad
$^2$Peking University
\quad
$^3$Meituan LongCat Team
\\
\texttt{{
tgui@fudan.edu.cn,
wye@pku.edu.cn,
wangwei432@meituan.com
}}
}

\begin{document}
\maketitle

\begin{abstract}
Recent advances in block diffusion language models have demonstrated competitive performance and strong scalability on reasoning tasks.
However, their test-time compute allocation remains largely unexplored, leaving a critical speed-effectiveness trade-off unresolved in long Chain-of-Thought reasoning. To address this, we propose a unified test-time compute allocation framework that introduces adaptivity in both step-wise decoding and blockwise generation.
At the decoding level, we propose Bounded Adaptive Confidence Decoding (BACD), a difficulty-aware sampling strategy that dynamically adjusts denoising based on model confidence, accelerating inference while controlling error accumulation.
Beyond step-wise adaptivity, we introduce the Think Coarse, Critic Fine (TCCF) paradigm that allocates large block sizes for exploratory thinking and smaller block sizes for precise refinement.
To stabilize training under varying block configurations, we adopt Progressive Block Size Extension, which mitigates quality degradation when scaling up block sizes.
Extensive evaluations on six benchmarks show that our TDAR-8B model with BACD and TCCF achieves a 2.38$\times$ speedup and +3.4\% average accuracy over the strong TraDo-8B baseline, unlocking the potential of block diffusion in complex reasoning.

\end{abstract}

\input{sections/introduction}

\input{sections/method}

\input{sections/experiment}

\input{sections/ablation}

\input{sections/related_work}
\input{sections/conclusion}

\input{sections/impact_statement}

\nocite{langley00}

\bibliography{example_paper}

\newpage
\appendix

\input{sections/appendix}

\end{document}

%% file: sections/introduction.tex
\section{Introduction}

Recent advances in Diffusion Large Language Models (DLLMs), exemplified by LLaDA~\citep{nie2025largellada}, Dream~\citep{ye2025dream}, Gemini Diffusion~\citep{genimidllm}, have demonstrated the scalability of this paradigm~\citep{nie2024scaling,gong2025scalingdiffusionlanguagemodels} and have achieved remarkable performance on mathematics~\citep{ye2025dream,cheng2025sdar}, code generation~\citep{gong2025diffucoderunderstandingimprovingmasked, wang2025revolutionizingtracerl}, and general reasoning tasks~\citep{ye2025autoregressiondiscretediffusioncomplex, shaorensijian2025diffuse}.
Building on the recent success of DLLMs, Block Diffusion Language Models (BDLMs)~\citep{arriola2025block} have emerged as a promising direction that integrates diffusion with traditional autoregressive (AR) decoding, enabling efficient KV caching and improving inference efficiency and generation flexibility.

\begin{figure}[t]
  \centering
  \includegraphics[width=1.0\columnwidth]{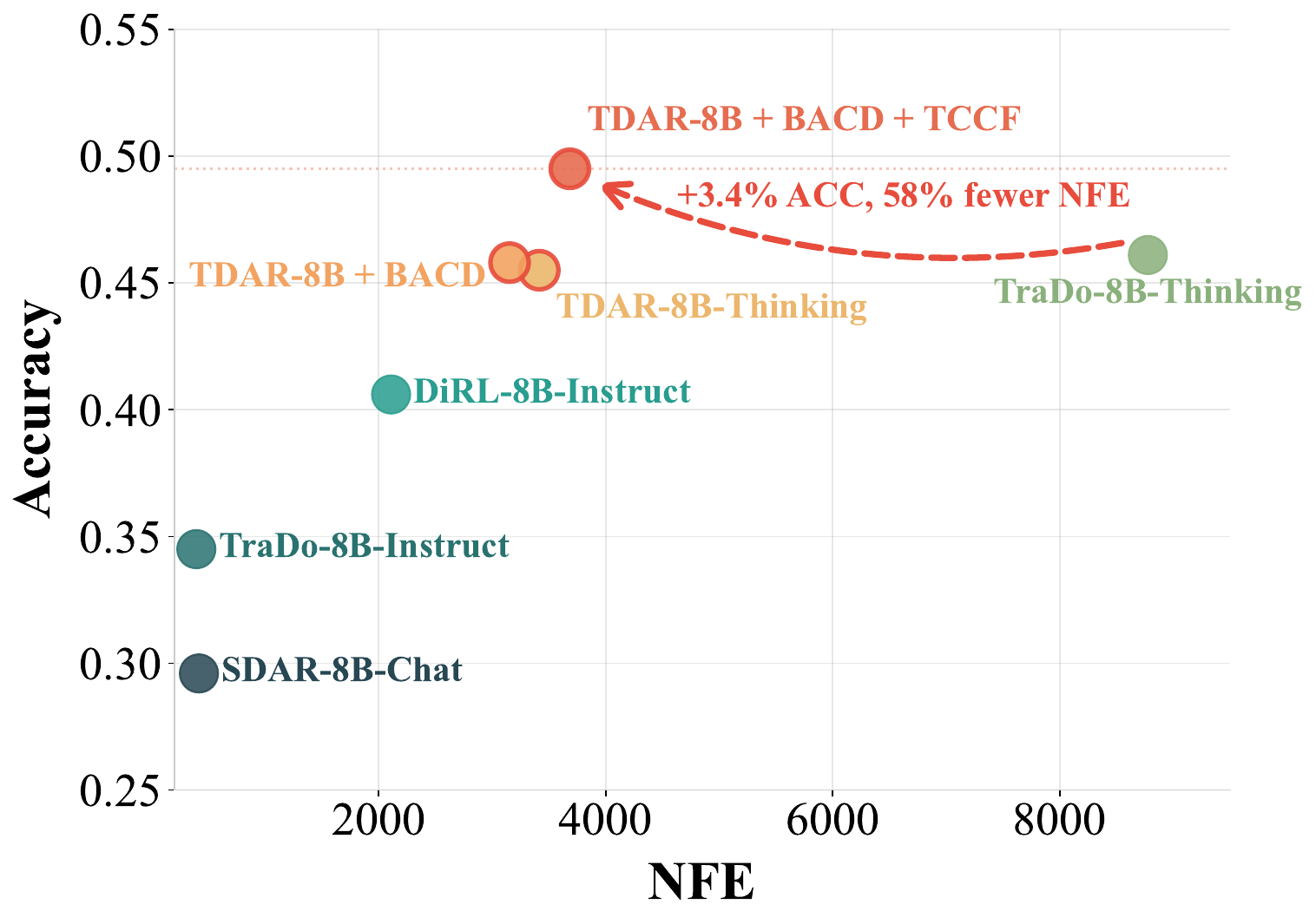}
  \caption{Performance and speed comparison of BDLMs. 
  Our TDAR-8B-Thinking achieves $2.38\times$ speedup and +3.4\% accuracy with BACD and TCCF compared to the best baselines.
  }
  \label{fig:aime24_performance_comparison}
  \vspace{-.5em}
\end{figure}

However, existing BDLMs~\citep{bie2025llada2, wang2025revolutionizingtracerl, zhu2025dirl} primarily focus on simple reasoning tasks and offer limited exploration 
under the test-time scaling paradigm. 
While \citet{wang2025revolutionizingtracerl} train a block diffusion model for complex reasoning, they do not systematically investigate how block diffusion should balance efficiency and effectiveness in such settings.
Test-time scaling with long Chain-of-Thought (CoT) reasoning presents a double-edged scenario for block diffusion, i.e., although it naturally favors parallel decoding, the increased reasoning complexity substantially intensifies the trade-off between efficiency and effectiveness.
In Figure \ref{fig:aime24_performance_comparison}, we observe that current BDLMs struggle to simultaneously achieve high performance and efficiency on complex reasoning tasks, where improvements in efficiency often lead to a substantial degradation in performance.


To address the efficiency–effectiveness trade-off of block diffusion under test-time compute allocation,
we propose a unified framework that introduces adaptivity in both decoding and generation strategies.
Our key insight is that long reasoning trajectories are inherently heterogeneous:
different parts vary not only in local sampling difficulty, but also in their role within the overall reasoning process.
Effectively exploiting this non-uniformity is crucial for achieving both efficiency and reliability in block diffusion.
At the decoding stage, we introduce \textbf{Bounded Adaptive Confidence Decoding (BACD)},
a dynamic sampling strategy that adapts the denoising process to the varying difficulty of long reasoning trajectories.
Using the average confidence of previously decoded tokens as a difficulty signal,
our method enables aggressive acceleration when the model is highly confident via an upper-bound threshold,
while enforcing a lower-bound threshold to prevent excessive error accumulation under uncertainty.
Beyond step-wise decoding adaptivity, we propose \textbf{Think Coarse, Critic Fine (TCCF)},
a test-time scaling paradigm that exploits the heterogeneous roles within long reasoning trajectories.
While exploratory reasoning can be generated efficiently with coarse-grained decoding,
refinement and summarization demand finer granularity to ensure correctness.
TCCF therefore allocates larger block sizes to exploratory segments
and smaller block sizes to refinement-oriented segments,
enabling an effective balance between efficiency and reasoning quality under test-time scaling.
To enable efficient and effective decoding in a large block size,
we adopt Progressive Block Size Extension, a multi-stage denoising fine-tuning strategy, 
which mitigates performance degradation when scaling block sizes.
These designs enable BDLMs to effectively balance efficiency and reasoning quality,
unlocking their potential for test-time scaling on complex reasoning tasks.

To evaluate the effectiveness and generalizability of our approach, we conduct extensive experiments on a diverse set of six reasoning benchmarks, including mathematics, code generation, and STEM. 
Results show that TDAR-8B-Thinking with BACD achieves state-of-the-art performance among 8B-scale block diffusion language models, while providing up to 4.14$\times$ speedup over standard autoregressive decoding. 
In addition, the TCCF paradigm further improves reasoning performance and offers a better trade-off between speed and accuracy. 
In summary, our contributions are threefold:
\begin{itemize}
    \item We propose Bounded Adaptive Confidence Decoding, a difficulty-aware decoding strategy that dynamically adapts denoising at test time, improving both performance and efficiency with strong robustness.
    \item We introduce Think Coarse, Critic Fine, a new test-time compute allocation paradigm that combines coarse-grained exploration with fine-grained refinement, significantly enhancing complex reasoning performance with minimal efficiency overhead.
    \item We identify block sizes as a key factor for BDLM acceleration and propose Progressive Block Size Extension, which mitigates performance degradation when scaling block sizes and unlocks BDLMs' parallel potential.
\end{itemize}

%% file: sections/method.tex
\section{Preliminary}
\label{sec:preliminary}
\subsection{Block Diffusion Language Models}
Block Diffusion Language Models~\citep{arriola2025block, wu2025fastdllmv2, cheng2025sdar} interpolate between autoregressive and diffusion paradigms. 
A token sequence $\mathbf{x}$ is partitioned into $K$ non-overlapping blocks $\mathbf{x} = (\mathbf{x}^1, \dots, \mathbf{x}^K)$, where the block size of $\mathbf{x}^k$ represents as $B$. The model factorizes the likelihood autoregressively over blocks:
\begin{equation}
\log p_\theta(\mathbf{x}) = \sum_{k=1}^{K} \log p_\theta(\mathbf{x}^k \mid \mathbf{x}^{<k}),
\end{equation}
where $\mathbf{x}^{<k}$ denotes the preceding clean blocks. 
Each conditional distribution $p_\theta(\mathbf{x}^k \mid \mathbf{x}^{<k})$ is modeled via a discrete masked diffusion process: a forward process and a reverse process. The forward process $q(\mathbf{x}^k_t \mid \mathbf{x}^k_0)$ gradually masks tokens within block $\mathbf{x}^k$. When adopt a linear noise schedule defined by $\alpha_t = 1-t$ at time $t \in [0, 1]$, each token is independently replaced by a special \texttt{[MASK]} token $\mathbf{m}$ with probability $1-\alpha_t$:
\begin{equation}
q(\mathbf{x}^k_t \mid \mathbf{x}^k_0) = \prod_{i=1}^{B} \text{Cat}((\mathbf{x}^k_t)_i; (1-t) (\mathbf{x}^k_0)_i + t \mathbf{m}).
\end{equation}
The reverse process reconstructs the original block $\mathbf{x}^k_0$ from the noisy state $\mathbf{x}^k_t$ given the history $\mathbf{x}^{<k}$. The model predicts the distribution of original tokens in parallel:
\begin{equation}
p_\theta(\mathbf{x}^k_0 \mid \mathbf{x}^k_t, \mathbf{x}^{<k}) = \prod_{i=1}^{B} p_\theta((\mathbf{x}^k_0)_i \mid \mathbf{x}^k_t, \mathbf{x}^{<k}).
\end{equation}


Standard block discrete denoising diffusion models minimize the Negative Evidence Lower Bound (NELBO). Under the linear noise schedule $\alpha_t = 1-t$, the objective is scaled by the inverse noise level $1/t$. The loss is computed exclusively on the set of masked positions $\mathcal{M}_t^k$ within each block:
\begin{equation}
\label{eq:NELBO}
\scalebox{0.95}{$
\displaystyle
\mathcal{L}(\theta) = \mathbb{E}_{t, \mathbf{x}} \left[
-\frac{1}{t}
\sum_{k=1}^{K}
\sum_{i \in \mathcal{M}_t^k}
\log p_\theta\big( (\mathbf{x}^k_0)_i \mid \mathbf{x}^k_t, \mathbf{x}^{<k} \big)
\right]
$}
\end{equation}

\subsection{Sampling Algorithm for BDLMs}
\label{sec:sampling_algorithm}
Generation in BDLMs proceeds block-by-block. For the $k$-th block $\mathbf{x}^k$, decoding starts from a fully masked state $\mathbf{x}^k_T$. At each denoising step $t$, the model predicts token probabilities for all masked positions $i \in \mathcal{M}_t^k$ conditioned on the current state $\mathbf{x}^k_t$, history $\mathbf{x}^{<k}$, and prompt $\mathbf{p}$. We define the \textit{confidence} score for position $i$ as:
\begin{equation}
c_i(\mathbf{x}^k_t) := \max_{v \in \mathcal{V}} p_\theta((\mathbf{x}^k_0)_i = v \mid \mathbf{x}^k_t, \mathbf{x}^{<k}, \mathbf{p}).
\end{equation}
A subset of masked tokens $\mathcal{D}_t \subseteq \mathcal{M}_t^k$ is then selected for unmasking based on $c_i$. We introduce two widely used strategies for BDLMs:

\paragraph{Static Confidence Decoding.~\citep{nie2025largellada}}
A fixed number of tokens, $n = \lceil B/N \rceil$, are decoded at each step, where $N$ is the total denoising steps. Specifically, $\mathcal{D}_t$ consists of the $n$ masked positions with the highest confidence scores. 

\paragraph{Dynamic Confidence Decoding.~\citep{wu2025fastdllm}}
Tokens exceeding confidence threshold $\tau$ are decoded, with the selection set defined as $\mathcal{D}_t = \{ i \in \mathcal{M}_t^k \mid c_i(\mathbf{x}^k_t) > \tau \}$. This accelerates inference by rapidly decoding easy tokens.

\begin{figure*}[t]
  \centering
  \includegraphics[width=\textwidth]{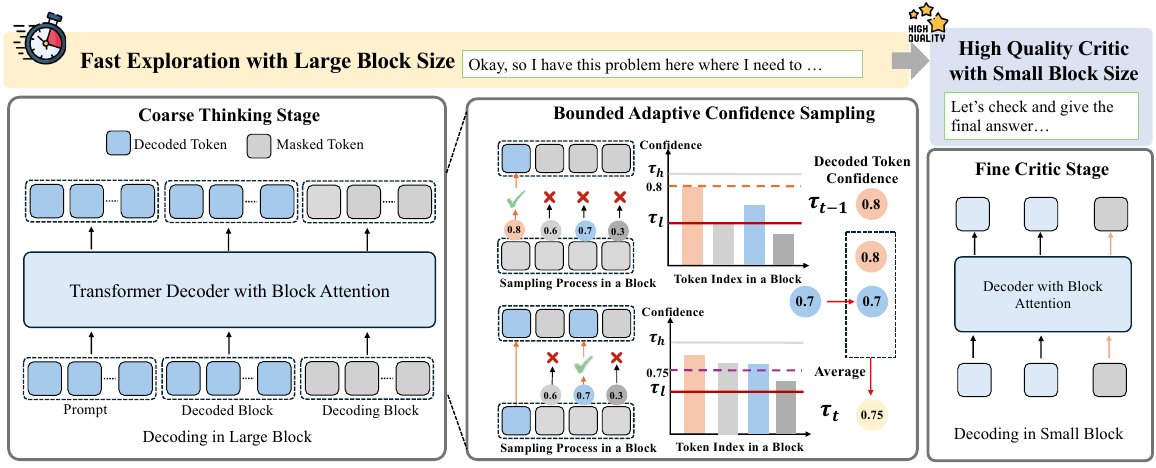}
\vspace{-20pt}
\caption{Overview of our reasoning process. We use Bounded Adaptive Confidence Decoding to enable fast exploration with large block sizes, and apply small block sizes for fine-grained refinement.}
  \label{fig:main_fig}
\end{figure*}

\section{Method}
We propose a unified framework for test-time scaling in Block Diffusion Language Models (BDLMs),
which exploits the non-uniform nature of long reasoning trajectories through adaptivity at two levels.
At the decoding level, we design a dynamic sampling strategy that adjusts denoising steps to local difficulty.
At the reasoning-phase level, we introduce a new test-time scaling paradigm that adapts block sizes
to different stages of reasoning.
An overview of our framework is shown in Figure~\ref{fig:main_fig}.

\subsection{Bounded Adaptive Confidence Decoding}
\label{sec:decoding_algorithm}

Dynamic decoding for block diffusion models typically relies on a fixed confidence threshold,
which becomes increasingly brittle as block size grows:
high thresholds lead to inefficient decoding,
while low thresholds cause error accumulation due to overly aggressive unmasking.
A fundamental reason is that the confidence distribution within a block is non-stationary: it
starts low when context is sparse and gradually rises as more tokens are unmasked.

To address this, we propose \textbf{Bounded Adaptive Confidence Decoding (BACD)}.
Instead of using a fixed threshold, BACD dynamically adjusts the acceptance threshold based on the running mean of confidence scores from previously accepted tokens, allowing the criterion to naturally adapt as more tokens are revealed within a block.
To prevent instability, we impose explicit upper and lower bounds on the threshold.
The detailed decoding procedure is summarized in Algorithm~\ref{alg:bacd}.

BACD maintains a list of confidence scores from previously decoded tokens
and uses their average to guide future decoding decisions.
Let $\mathbf{c}_{t-1}$ denote the list of confidence scores collected before step $t$.
In the initial state where $\mathbf{c}_{t-1}$ is empty, we set $\tau_t = \tau_h$.
At each decoding step, we compute the average confidence $\bar{c}_{t-1} = \text{mean}(\mathbf{c}_{t-1})$.
To stabilize decoding behavior, we introduce two thresholds:
an upper bound $\tau_h$ and a lower bound $\tau_l$.
The effective decoding threshold $\tau_t$ is determined via a clipping operation:
\begin{equation}
\tau_t = \text{clip}(\bar{c}_{t-1}, \tau_l, \tau_h).
\end{equation}

Using $\tau_t$, we select the set of tokens to be unmasked at step $t$:
\begin{equation}
\mathcal{D}_t =
\{ i \in \mathcal{M}_t^k \mid c_i(\mathbf{x}^k_t) > \tau_t \}.
\end{equation}
To guarantee convergence, if $\mathcal{D}_t = \varnothing$,
we unmask the token with the highest confidence score.
The confidence scores of newly decoded tokens are then appended
to the list $\mathbf{c}$ for subsequent steps.
By bounding the adaptive threshold, BACD prevents overly conservative
or overly aggressive decoding, enabling a more stable trade-off
between efficiency and generation quality.

\begin{algorithm}[ht]
\caption{Bounded Adaptive Confidence Decoding}
\label{alg:bacd}
\KwIn{Masked block $\mathbf{x}_{T}^k$, bounds $\tau_h, \tau_l$}
\KwOut{Clean block $\mathbf{x}_{0}^k$}

Initialize confidence list $\mathbf{c} \leftarrow [\ ]$\;
\While{masked tokens exist in $\mathbf{x}^k$}{
    Let $\mathcal{M} \leftarrow \{ i \mid (\mathbf{x}^k)_i = \texttt{[MASK]} \}$\;
    Compute confidence scores $c_i$ for all $i \in \mathcal{M}$\;
    
    Calculate mean $\bar{c} \leftarrow \text{mean}(\mathbf{c})$ if $\mathbf{c} \neq [\ ]$ else $\tau_h$\;
    Set threshold $\tau \leftarrow \text{clip}(\bar{c}, \tau_l, \tau_h)$\;
    
    Select candidates $\mathcal{D} \leftarrow \{ i \in \mathcal{M} \mid c_i > \tau \}$\;
    \lIf{$\mathcal{D} = \varnothing$}{ $\mathcal{D} \leftarrow \{\arg\max_{i \in \mathcal{M}} c_i\}$ }

    Unmask tokens at indices $\mathcal{D}$, append scores to $\mathbf{c}$\;
}
\end{algorithm}
\vspace{-0.5em}

\subsection{Think Coarse, Critic Fine}
\label{sec:our_method}

Long CoT reasoning is not uniform across its trajectory.
Different segments of a reasoning process serve distinct functional roles.
In particular, early segments often focus on exploratory reasoning,
while later segments emphasize refinement, verification, and summarization.
However, block diffusion models exhibit a clear trade-off between block size and generation quality:
larger block sizes enable faster decoding but may degrade accuracy,
whereas smaller block sizes improve precision at the cost of efficiency.
As a result, a single block size is insufficient to balance efficiency and reasoning quality across all stages.

Motivated by this observation, we propose \textbf{Think Coarse, Critic Fine (TCCF)},
a test-time scaling paradigm that adapts the block size to different stages of reasoning.
TCCF leverages the trade-off between block size and generation quality in block diffusion,
allocating computation according to the functional role of each reasoning segment.
Given an input prompt $\mathbf{p}$, TCCF decomposes inference into two stages:

\paragraph{Coarse Thinking.}
The model first generates an exploratory reasoning trajectory $\mathbf{r}$
using a large block size $B_{\text{think}}$ to maximize efficiency:
\begin{equation}
\mathbf{r} \sim p_\theta(\cdot \mid \mathbf{p}; B_{\text{think}}).
\end{equation}

\paragraph{Fine Critic.}
Conditioned on $\mathbf{p}$ and the exploratory trajectory $\mathbf{r}$,
the model performs refinement and consolidation using a smaller block size
$B_{\text{critic}} < B_{\text{think}}$ to improve reasoning reliability:
\begin{equation}
\mathbf{y} \sim p_\theta(\cdot \mid \mathbf{p}, \mathbf{r}; B_{\text{critic}}).
\end{equation}

\paragraph{Training with Large Block Sizes.}
TCCF relies on large block sizes during coarse thinking,
which can lead to performance degradation if the model is trained
with a fixed small block size.
To enable stable training under large block sizes,
we adopt \textbf{Progressive Block Size Extension},
a multi-stage supervised fine-tuning strategy that gradually increases the block size.
Given a training instance $(\mathbf{p}, \mathbf{x})$ consisting of a prompt $\mathbf{p}$ and a target response $\mathbf{x}$,
the supervised fine-tuning objective is:
\begin{equation}
\label{eq:sft_loss}
\scalebox{0.85}{$
\displaystyle
\mathcal{L}_{\text{SFT}}(\theta) =
\mathbb{E}_{t, (\mathbf{p}, \mathbf{x})}
\left[
-\frac{1}{t}
\sum_{k=1}^{K}
\sum_{i \in \mathcal{M}_t^k}
\log p_\theta\big(
(\mathbf{x}^k_0)_i \mid \mathbf{x}^k_t, \mathbf{x}^{<k}, \mathbf{p}
\big)
\right].
$}
\end{equation}
This progressive scaling mitigates performance degradation during expansion, which is crucial for enabling TCCF's coarse thinking stage.

%% file: sections/experiment.tex
\section{Experiment}

\begin{table*}[ht]
    \centering
    \vspace{-0.5em}
    \definecolor{oursBlue}{RGB}{0, 120, 215}
    \definecolor{RoyalBlue}{RGB}{65, 105, 225}
    \renewcommand\arraystretch{1.05}
    \resizebox{\textwidth}{!}{
\begin{tabular}{@{}lccccccccccccccc@{}}
    \toprule
    \multirow{2}{*}{\textbf{Method}} & \multicolumn{2}{c}{\textbf{Math500}} & \multicolumn{2}{c}{\textbf{AIME24}} & \multicolumn{2}{c}{\textbf{AIME25}} & \multicolumn{2}{c}{\textbf{AMC23}} & \multicolumn{2}{c}{\textbf{LCB}} & \multicolumn{2}{c}{\textbf{GPQA}} & \multicolumn{3}{c}{\textbf{AVG}} \\
    \cmidrule(lr){2-3} \cmidrule(lr){4-5} \cmidrule(lr){6-7} \cmidrule(lr){8-9} \cmidrule(lr){10-11} \cmidrule(lr){12-13} \cmidrule(lr){14-16}
    & TPF & ACC & TPF & AVG@32 & TPF & AVG@32 & TPF & AVG@32 & TPF & AVG@8 & TPF & AVG@8 & TPF & NFE & ACC \\
    \midrule
    \rowcolor{gray!10} \multicolumn{16}{c}{\emph{Masked Diffusion Language Model}} \\
    \hspace{1.0mm}LLaDA & 3.92 & 41.2 & 3.44 & 6.7 & 3.66 & 0.0 & 4.07 & 12.5 & 2.83 & 4.7 & 2.91 & 17.2 & 3.47 & 233 & 13.7 \\
    \hspace{1.0mm}LLaDA-1.5 & 3.97 & 42.2 & 3.35 & 0.0 & 3.69 & 0.0 & 4.01 & 10.0 & 2.86 & 4.3 & 3.01 & 24.2 & 3.48 & 236 & 13.5 \\
    \hspace{1.0mm}LLaDA-MoE & 2.71 & 56.6 & 2.89 & 3.3 & 2.71 & 0.0 & 3.17 & 32.5 & 2.06 & 12.9 & 2.18 & 27.8 & 2.62 & 259 & 22.2 \\
    \midrule
    \rowcolor{gray!10} \multicolumn{16}{c}{\emph{Block Diffusion Language Model}} \\
    \hspace{1.0mm}Fast-dLLM-v2 & 2.81 & 59.4 & 2.59 & 0.0 & 2.58 & 0.0 & 2.78 & 25.0 & 1.73 & 6.8 & 2.09 & 28.3 & 2.43 & 259 & 19.9 \\
    \hspace{1.0mm}SDAR-8B-Chat & 2.59 & 71.2 & 3.01 & 7.0 & 2.33 & 9.5 & 2.63 & 45.3 & 2.32 & 13.7 & 1.98 & 30.9 & 2.48 & 416 & 29.6 \\
    \hspace{1.0mm}DiRL-8B-Instruct & 2.09 & 80.0 & 1.88 & 21.0 & 1.82 & 19.9 & 1.99 & 62.2 & 2.00 & 15.9 & 1.76 & 44.4 & 1.92 & 2109 & 40.6 \\
    \hspace{1.0mm}TraDo-8B-Instruct & 2.39 & 74.8 & 2.16 & 10.7 & 2.01 & 13.0 & 2.22 & 55.8 & 1.73 & 17.2 & 1.44 & 35.5 & 1.99 & 393 & 34.5 \\
    \hspace{1.0mm}TraDo-8B-Thinking & 1.42 & 85.8 & 1.32 & 30.6 & 1.32 & 24.5 & 1.36 & 69.7 & 1.22 & 22.2 & 1.30 & 44.0 & 1.32 & 8777 & 46.1 \\
    \hspace{1.0mm}SDAR-30B-Thinking & 1.66 & 91.6 & 1.63 & 61.6 & 1.60 & 53.6 & 1.62 & 91.8 & 1.31 & 38.2 & 1.37 & 64.9 & 1.53 & 7449 & 67.0 \\
    \midrule
    \hspace{1.0mm}TDAR-8B-Thinking (ours) & 3.98 & 78.0 & 3.76 & 35.7 & 3.50 & 29.8 & 4.00 & 68.8 & 2.67 & 30.6 & 4.64 & 30.2 & 3.76 & 3415 & 45.5 \\
    \rowcolor{oursBlue!12}
    \hspace{2.5mm}+ BACD & 4.61 & 78.8 & 4.05 & 36.1 & 3.79 & 30.5 & 4.43 & 69.9 & 2.96 & 29.4 & 4.98 & 30.2 & 4.14 & 3153 & 45.8 \\
    \rowcolor{RoyalBlue!30}
    \hspace{2.5mm}+ BACD +TCCF & 3.67 & 82.8 & 3.84 & 39.8 & 3.56 & 32.7 & 3.86 & 73.9 & 2.75 & 33.1 & 4.27 & 34.4 & 3.66 & 3684 & 49.5 \\
    \midrule
    \hspace{1.0mm}TDAR-30B-Thinking (ours) & 1.55 & 90.6 & 1.50 & 66.8 & 1.54  & 56.5 & 1.56 & 93.4 & 1.33 &  50.7 & 1.46 &  52.8 &  1.49 & 6402 & 68.5 \\
    \rowcolor{oursBlue!12}
    \hspace{2.5mm}+ BACD & 1.63 & 90.8 & 1.56 & 69.1 & 1.60 & 55.7 & 1.60 & 93.3 & 1.38 & 51.5 & 1.51 & 52.6 & 1.55 & 6128 & 68.8 \\
    \rowcolor{RoyalBlue!30}
    \hspace{2.5mm}+ BACD +TCCF & 1.62 & 91.0 & 1.57 & 69.1 & 1.60 & 57.4 & 1.62 & 94.7 & 1.39 & 51.4 & 1.52 & 56.8 & 1.55 & 6580 & 70.1 \\
    \bottomrule
\end{tabular}
    }
\caption{Performance comparison of various models on six reasoning benchmarks.}
\label{tab:main_table}
\end{table*}

\subsection{Experiment Setup}

\paragraph{Adaptation to Long CoT Reasoning BDLMs.}
We adapt Qwen3-8B-base into Block DLMs following~\citet{cheng2025sdar}. We first train with a blockwise diffusion objective ($B=4$) on a 50B-token annealing corpus, then perform SFT on long CoT data with progressive block-size expansion from $B=4$ to $B=64$.
Based on the performance-efficiency trade-off analysis in Section~\ref{sec:analysis_blocksize_extension}, we select the model trained with a block size of $B=16$ for 8B models and $B=4$ for 30B models as our reasoning model denoted as \textbf{TDAR-8B-thinking} and \textbf{TDAR-30B-thinking}.
Implementation details are provided in Appendix~\ref{appendix:implementation_training}.

\paragraph{Datasets and Baselines.}

To comprehensively evaluate BDLMs, we select representative benchmarks covering mathematical reasoning (Math500~\citep{hendrycks2021math500}, AIME2024~\citep{aime2024}, AIME2025~\citep{aime2025}, AMC2023~\citep{AMC}), code generation (LiveCodeBench v5\footnote{The evaluation time range spans from August 2024 to May 2025.}~\citep{jain2024livecodebench}), and STEM reasoning (GPQA-diamond~\citep{rein2023gpqagraduatelevelgoogleproofqa}).  
Specifically, we configure 32 runs for AIME2024, AIME2025, and AMC2023; 8 runs for LiveCodeBench and GPQA-diamond.

We compare our method against state-of-the-art open-source BDLMs, including Fast-dllm-v2~\citep{wu2025fastdllmv2}, SDAR-8B-Chat~\citep{cheng2025sdar}, SDAR-30B-Thinking~\citep{cheng2025sdar}, DiRL-8B-Instruct~\citep{zhu2025dirl}, TraDo-8B-Instruct~\citep{wang2025revolutionizingtracerl}, and TraDo-8B-Thinking~\citep{wang2025revolutionizingtracerl}.
To compare with masked diffusion language models, we involve LLaDA~\citep{nie2025largellada}, LLaDA-1.5~\citep{zhu2025llada15variancereducedpreference}, LLaDA-MoE~\citep{zhu2025lladamoesparsemoediffusion}.

\paragraph{Evaluation Setup}
We adopt the block size $B$ recommended in their official repositories: $B=4$ for SDAR-8B-Chat, DiRL-8B-Instruct, and TraDo-8B; $B=32$ for Fast-dllm-v2.
For decoding algorithms, we adapt widely used algorithms from DLLMs: Dynamic Confidence-Aware Decoding~\citep{wu2025fastdllm}.
We employ Dynamic Confidence Decoding with a threshold $\tau=0.9$ as the default algorithm for the main results. 
For BACD, we set the upper bound $\tau_{h}$ to 0.9 and the lower bound $\tau_{l}$ to 0.7. 
Other details are shown in Appendix~\ref{appendix:implementation_inference}.
For the TCCF paradigm, we configure TDAR-8B-thinking with $B_{\text{think}}=16$ and $B_{\text{critic}}=4$ and TDAR-30B-thinking with $B_{\text{think}}=B_{\text{critic}}=4$. TCCF implementation details are provided in Appendix~\ref{appendix:detailed_usage_of_tccf}.

\paragraph{Evaluation Metrics of Efficiency.}

We evaluate computational efficiency using two key metrics: Tokens Per Forward Pass (TPF) and Number of Function Evaluations (NFE). 
While generating a complete reasoning sequence, NFE denotes the actual number of forward passes and TPF measures the average number of tokens generated per forward model call, defined as:
\begin{equation*}
    \text{TPF} = \frac{\text{Total Generated Tokens}}{\text{NFE}}
\end{equation*}
Higher TPF and lower NFE values indicate greater decoding parallelism and lower inference overhead. In our main evaluation, we report benchmark-specific TPF alongside the overall average TPF and NFE, with more exhaustive efficiency and throughput analyses provided in Appendix~\ref{averge_nfe_tok}.

\begin{figure*}[ht]
  \centering
  \includegraphics[width=\textwidth]{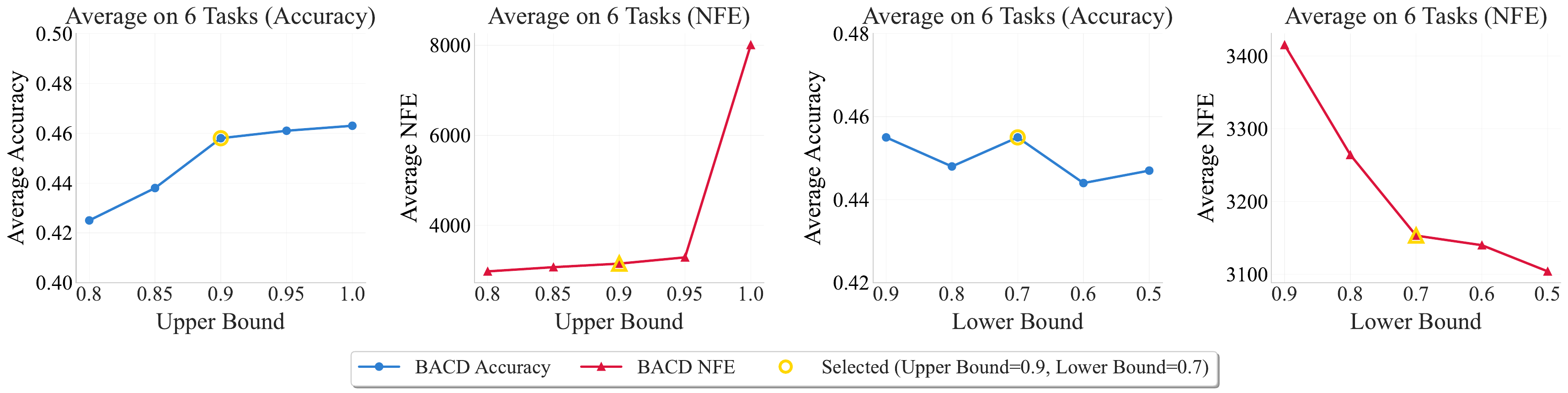}
  \caption{Accuracy and Speed under different thresholds across 6 tasks. 
  }
  \label{fig:bacd_threshold}
  \vspace{-.5em}
\end{figure*}

\subsection{Main Results}

\paragraph{Performance and Speed}
As shown in Table~\ref{tab:main_table}, TDAR-8B-thinking achieves competitive reasoning performance alongside outstanding decoding efficiency across a wide range of benchmarks. 
Specifically, our base model, TDAR-8B-thinking, achieves an average decoding speed of 3.76 TPF, which represents a \textbf{2.57$\times$} speedup in NFE compared to the strongest BDLM baseline, TraDo-8B-Thinking, while maintaining comparable accuracy. 
When fully equipped with our proposed BACD decoding algorithm and the TCCF paradigm, TDAR-8B-thinking achieves a final average accuracy of \textbf{49.5\%}. 
This configuration outperforms TraDo-8B-Thinking by exactly 3.4 percentage points while maintaining a highly efficient decoding speed of 3.66 TPF, delivering a \textbf{2.38$\times$} speedup in NFE.

\paragraph{Effectiveness of BACD and TCCF}

Integrating the BACD decoding algorithm and the TCCF paradigm further optimizes the speed-accuracy trade-off of TDAR-8B-thinking. 
Specifically, BACD dynamically accelerates the decoding process by reducing average NFE while maintaining or slightly improving overall generation accuracy. 
Meanwhile, the TCCF paradigm significantly enhances reasoning performance on complex tasks by allocating computing budgets adaptively. 
For instance, on the highly challenging AIME24 benchmark, TCCF improves the reasoning accuracy of TDAR-8B-thinking to \textbf{39.8\%} (a 3.7\% absolute gain over the BACD variant) while still preserving a high decoding speed of 3.84 TPF. 
Across all evaluations, the synergy of BACD and TCCF demonstrates that adaptive block sizing and confidence bounding can robustly balance inference cost and deep reasoning quality.

\paragraph{Generalization Across Model Scales}
We further evaluate the scalability and generalization of our approach by scaling up to a larger model size. 
As shown in Table~\ref{tab:main_table}, TDAR-30B-thinking achieves a base average accuracy of \textbf{68.5\%} with \textbf{1.49 TPF}, already outperforming the strong SDAR-30B-Thinking baseline in both accuracy and efficiency. 
Applying BACD to the 30B model increases the average speed to \textbf{1.55 TPF} while slightly boosting accuracy to \textbf{68.8\%}. 
Finally, equipped with TCCF, TDAR-30B-thinking reaches a strong average accuracy of \textbf{70.1\%}. This demonstrates that our training and decoding paradigms generalize robustly to larger scales.


%% file: sections/ablation.tex
\section{Analysis}

\begin{figure*}[ht]
    \centering
    \includegraphics[width=0.99\textwidth]{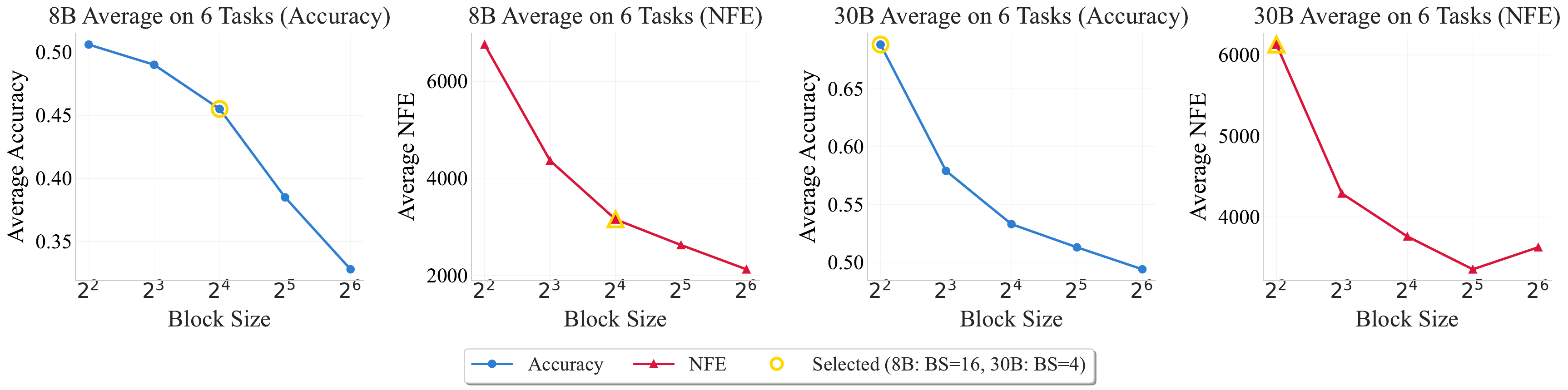}
    \caption{Impact of block size on 8B and 30B models' performance and efficiency. Gold marker indicates our selected checkpoint.}

    \label{fig:block_size_trade_off}

    \vspace{-.5em}

\end{figure*}

\subsection{Analysis on Sampling Algorithm}
\label{sec:analysis_decoding_algorithm}

\paragraph{Comparison with Existing Decoding Algorithms.}






We compare BACD ($\tau_h{=}0.9, \tau_l{=}0.7$) against mainstream decoding algorithms on TDAR-8B-Thinking. The baselines include: (1) \textbf{Static methods} (LowConfidenceStatic, EntropyStatic) with a fixed step budget of $N{=}B/4{=}4$; (2) \textbf{Dynamic Confidence}~\citep{wu2025fastdllm} (our default setting, with $\tau{=}0.9$); (3) \textbf{EB-Sampler}~\citep{benhamu2025acceleratedsamplingmaskeddiffusion} using confidence, entropy, and margin proxies ($\gamma{=}0.5$); and (4) \textbf{Saber}~\citep{dong2025saber} ($\mu{=}2$). Detailed configurations for each baseline are provided in Appendix~\ref{appendix:decoding_details}.
    
\textbf{Results.}
As shown in Table~\ref{tab:decoding_comparison}, static methods suffer from catastrophic accuracy collapse, confirming that fixed decoding schedules are unsuitable for BDLMs. Among dynamic methods, EB-Sampler (entropy) achieves the highest accuracy (48.1\%) but incurs severe computational overhead (7,358 NFE). Saber yields suboptimal accuracy despite high NFE. In contrast, BACD achieves the best overall trade-off, outperforming the default Dynamic Confidence baseline in both accuracy (+0.3\%) and efficiency (reducing NFE to 3,153, a 7.7\% reduction), while delivering the highest throughput (4.14 TPF) among all baselines.

\begin{table}[ht]
\centering
\resizebox{\columnwidth}{!}{
\begin{tabular}{lccc}
\toprule
Method & Acc & NFE & TPF \\
\midrule
StaticConfidence & 17.8 & 3707 & 4.00 \\
StaticEntropy & 3.5 & 3535 & 4.00 \\
\midrule
DynamicConfidence  & 45.5 & 3415 & 3.76 \\
EB-Sampler (confidence) & 44.9 & 3353 & 3.66 \\
EB-Sampler (entropy) & \textbf{48.1} & 7358 & 1.55 \\
EB-Sampler (margin) & 43.3 & 7677 & 1.56 \\
Saber & 34.9 & 3977 & 2.95 \\
\midrule
BACD & 45.8 & \textbf{3153} & \textbf{4.14} \\
\bottomrule
\end{tabular}
}
\caption{Comparison of BACD with existing decoding algorithms on TDAR-8B-Thinking, averaged over six benchmarks.}
\label{tab:decoding_comparison}
\end{table}

\paragraph{Performance and Speed under Different Thresholds.}
We ablate the impact of varying upper bound $\tau_h$ and lower bound $\tau_l$ in BACD across all 6 tasks in Figure~\ref{fig:bacd_threshold}. As $\tau_h$ increases from 0.8 to 0.95, accuracy steadily improves with only a modest NFE increase, since a higher upper bound makes the clipping operation more conservative, allowing fewer tokens to pass the threshold per step and thus improving generation quality. However, setting $\tau_h = 1.0$ causes the NFE to surge dramatically, because it reaches the theoretical ceiling of confidence scores, forcing the algorithm to repeatedly fall back to the single-token fallback rule ($\mathcal{D} \leftarrow \{\arg\max_{i} c_i\}$) and degenerating toward token-by-token decoding. For $\tau_l$, varying it from 0.9 down to 0.5 yields a slight NFE reduction while accuracy degrades gradually, as a lower bound permits more aggressive early unmasking when the adaptive mean $\bar{c}$ drops, accepting lower-confidence tokens that may introduce errors. We select $\tau_h=0.9$ and $\tau_l=0.7$ as the final configuration, which achieves competitive accuracy (45.8\%) at a moderate NFE (3153) while avoiding the efficiency collapse at extreme upper bounds.

\subsection{Analysis on Block Size Extension}

\label{sec:analysis_blocksize_extension}


Larger block sizes accelerate inference by reducing NFE but degrade generation quality due to the challenges of parallel prediction. As the block size $B$ grows, approximating $p(x_{t+1}, \ldots, x_{t+B} \mid x_{\leq t})$ in a single forward pass diverges further from the true autoregressive factorization, compounding prediction errors, as Figure~\ref{fig:block_size_trade_off} shown. 

To balance this trade-off, we select the block size at the knee point of the Accuracy--NFE curve (details in Appendix~\ref{app:block_size_knee_point}), where the marginal accuracy loss per unit NFE reduction sharply increases. For the 8B model, the marginal cost (accuracy loss per 1,000 NFE saved) rises from 0.029 ($B=8\to16$) to 0.133 ($B=16\to32$), a 4.6$\times$ increase that identifies $B=16$ as the optimal knee point. A similar analysis identifies $B=4$ for the 30B model, leading us to adopt $B=16$ for the 8B model and $B=4$ for the 30B model.



\section{Ablation Study}

\subsection{Ablation on Progressive Block Size Extension}

We conduct an ablation study to evaluate the effectiveness of our progressive block-size expansion strategy compared to directly training with a large fixed block size. We train two variants of the TDAR-8B-Base model with a target block size of $B=16$ for an equivalent number of tokens.

As shown in Table~\ref{tab:ablation_progressive}, the progressive strategy yields superior results across all benchmarks, suggesting that it effectively mitigates the optimization difficulties associated with large block sizes. On average across six tasks, the progressive approach outperforms the direct baseline by 6.7 points in accuracy, while also reducing NFE by 213.9, indicating that progressive training leads to both better performance and more efficient inference.

\begin{table}[h]
\centering
\begin{small}
\resizebox{\columnwidth}{!}{
\begin{tabular}{lccc}
\toprule
Strategy & Acc & NFE & TPF \\
\midrule
Direct Expansion  & 39.1 & 3366.9 & \textbf{4.40} \\
\textbf{Progressive Expansion} & \textbf{45.8} & \textbf{3153.0} & 4.14 \\
\bottomrule
\end{tabular}
}
\end{small}
\caption{Ablation on block size expansion strategies. }
\label{tab:ablation_progressive}
\end{table}

\subsection{Ablation on Think Coarse Critic Fine Paradigm}


We conduct an ablation study on the block sizes used in the thinking stage ($B_{\text{think}}$) and the critic stage ($B_{\text{critic}}$) of the TCCF paradigm, with results shown in Table~\ref{tab:ablation_2stage_blocksize}. Reducing $B_{\text{critic}}$ consistently enhances reasoning accuracy, supporting our hypothesis: while the exploratory phase benefits from larger blocks to quickly navigate the search space, the final answer-formulation phase requires fine-grained prediction to prevent cascading errors.

However, accuracy is not always monotonic. For TDAR-8B (trained progressively up to $B=16$), performance peaks at $B_{\text{critic}}=4$ and declines at $B_{\text{critic}}=1$. This suggests that abrupt single-token decoding creates a distribution mismatch with the blockwise training objective; thus, $B_{\text{critic}}=4$ acts as a ``sweet spot'' balancing block-level coherence with precision. Conversely, TDAR-30B shows monotonic gains down to $B_{\text{critic}}=1$ due to its smaller base size ($B_{\text{think}}=4$). Crucially, this transition incurs minimal efficiency overhead: reducing TDAR-8B's $B_{\text{critic}}$ from 16 to 4 yields a substantial 1.5-point accuracy gain with under 7\% NFE overhead, confirming TCCF's Pareto-optimal efficiency in concentrating compute on the critique phase.

\begin{table}[ht]
    \centering
    
    \renewcommand{\arraystretch}{1.15}
    \setlength\dashlinedash{3pt}
    \setlength\dashlinegap{2pt}
    \setlength\arrayrulewidth{0.4pt}
    \resizebox{\columnwidth}{!}{
    \begin{tabular}{l cc ccc}
        \toprule
        \textbf{Model} & $\mathbf{B_{\text{think}}}$ & $\mathbf{B_{\text{critic}}}$ & \textbf{Acc} & \textbf{NFE} & \textbf{TPF} \\
        \midrule
        TDAR-8B  & 16 &  1 & 48.9 & 4251.2 & 3.15 \\
        TDAR-8B  & 16 &  2 & 49.5 & 3930.9 & 3.42 \\
        TDAR-8B  & 16 &  4 & 49.5 & 3684.4 & 3.66 \\
        TDAR-8B  & 16 &  8 & 48.3 & 3531.6 & 3.83 \\
        TDAR-8B  & 16 & 16 & 48.0 & 3453.9 & 3.94 \\
        \midrule
        TDAR-30B & 4 &  4 & 70.1 & 6579.7 & 1.55 \\
        TDAR-30B & 4 &  2 & 70.1 & 6730.3 & 1.51 \\
        TDAR-30B & 4 &  1 & 70.3 & 6982.4 & 1.45 \\
        \bottomrule
    \end{tabular}
    }
    \caption{Ablation on TCCF paradigm with different block sizes, averaged over six benchmarks.}
    \label{tab:ablation_2stage_blocksize}
\end{table}

%% file: sections/related_work.tex
\section{Related Work}

\subsection{Test Time Scaling and Efficient Reasoning}
The success of OpenAI's o1 popularized test-time compute scaling, which improves reasoning performance by scaling up inference-time computation~\citep{openai2024openaio1card}. However, this paradigm often leads to increasingly lengthy reasoning trajectories and ``overthinking'' behaviors~\citep{chen2025think23overthinkingo1like}. Existing solutions primarily focus on controlling generation length via length-controlled policy optimization~\citep{aggarwal2025l1controllinglongreasoning} or adaptive fine-tuning based on task complexity~\citep{hao2024traininglargelanguagemodels,liu2025learnreasonefficientlyadaptive,fang2025thinklessllmlearnsthink,arora2025traininglanguagemodelsreason,zhang2025adaptthinkreasoningmodelslearn}. In contrast, we explore leveraging Block Diffusion Language Models (BDLMs) to achieve parallel decoding and flexible sequence manipulation, offering a hardware-friendly pathway to accelerate test-time scaling.

\subsection{Diffusion Language Models}
Diffusion Language Models (DLMs)~\citep{nie2025largellada,ye2025dream,austin2023structureddenoisingdiffusionmodels,sahoo2024simpleeffectivemaskeddiffusion} present a promising alternative to sequential generation by enabling parallel multi-token predictions. To support efficient Key-Value (KV) caching in bidirectional setups, BDLMs~\citep{arriola2025block} employ a blockwise generation structure. Recent works adapt pre-trained language models to diffusion objectives~\citep{gong2025scalingdiffusionlanguagemodels, ye2025dream,wu2025fastdllmv2, cheng2025sdar} and incorporate reinforcement learning to enhance reasoning capabilities~\citep{wang2025revolutionizingtracerl,zhu2025dirl}.
However, BDLMs suffer from an inherent trade-off between parallelizability and generation quality. While several advanced dynamic samplers have been developed for masked diffusion models~\citep{benhamu2025acceleratedsamplingmaskeddiffusion,hong2025wideinnarrowoutrevokabledecoding,dong2025saber}, existing block diffusion baselines typically rely on static step budgets or constant confidence thresholds~\citep{cheng2025sdar,bie2025llada2}.

%% file: sections/conclusion.tex
\section{Conclusion}


In this paper, we propose a unified test-time compute allocation framework for BDLMs that adaptively balances reasoning quality and decoding efficiency. By introducing Bounded Adaptive Confidence Decoding (BACD) and the Think Coarse, Critic Fine (TCCF) paradigm, our approach dynamically adjusts step-level confidence bounds and block-level reasoning stages. Extensive evaluations across six benchmarks demonstrate that our framework significantly accelerates inference while improving reasoning accuracy. Overall, this work advances the efficiency-effectiveness Pareto frontier for block diffusion language models and lays a solid foundation for future research.

%% file: sections/impact_statement.tex

\section*{Limitations}
\label{sec:limitations}

A primary bottleneck appears in block-size scalability as model parameters increase; although the 8B model effectively preserves reasoning capabilities up to a block size of $B=16$, the larger 30B model experiences a steeper accuracy decline beyond $B=4$ . This suggests that simultaneously predicting high-dimensional joint token distributions becomes exponentially more challenging as model capacity expands, presenting an open question for scaling blockwise predictions. Beyond these scaling constraints, our efficiency evaluations primarily focus on algorithmic metrics. In practical deployment, dynamically adjusting block sizes during inference requires real-time memory layout transformations and KV Cache reallocation. Although modern inference engines can handle these adjustments, translating our theoretical speedups into optimal wall-clock latency in production environments warrants further research into system-level I/O and custom GPU kernel optimizations.

%% file: sections/appendix.tex
\section{Average Token Length and NFE of Different Models and Decoding Methods}
\label{averge_nfe_tok}

Table~\ref{tab:nfe_tokens} reports the average number of function evaluations (NFE) and average response token length (Tok) across six reasoning benchmarks. 
NFE reflects the primary computational cost of generation, while Tok indicates the verbosity of model outputs. 
Notably, the two-stage decoding (TCCF) paradigm moderately increases the NFE compared to its single-stage BACD counterpart (e.g., rising from 3,153 to 3,684 for TDAR-8B). 
This computational investment represents a deliberate test-time scaling behavior: by dedicating a slightly higher NFE budget to the fine-grained critic stage, the model achieves significantly enhanced output accuracy. 
This trade-off demonstrates that concentrating the computational budget specifically on the final verification and answer-formulation stage is a highly effective strategy for complex reasoning tasks.

\begin{table*}[ht]
    \centering
    
    \vspace{-0.5em}
    
    \definecolor{oursBlue}{RGB}{0, 120, 215}
    \definecolor{RoyalBlue}{RGB}{65, 105, 225}
    \renewcommand\arraystretch{1.05}
    \resizebox{\textwidth}{!}{
\begin{tabular}{@{}lcccccccccccccc@{}}
    \toprule
    \multirow{2}{*}{\textbf{Method}} & \multicolumn{2}{c}{\textbf{Math500}} & \multicolumn{2}{c}{\textbf{AIME24}} & \multicolumn{2}{c}{\textbf{AIME25}} & \multicolumn{2}{c}{\textbf{AMC23}} & \multicolumn{2}{c}{\textbf{LCB}} & \multicolumn{2}{c}{\textbf{GPQA}} & \multicolumn{2}{c}{\textbf{AVG}} \\
    \cmidrule(lr){2-3} \cmidrule(lr){4-5} \cmidrule(lr){6-7} \cmidrule(lr){8-9} \cmidrule(lr){10-11} \cmidrule(lr){12-13} \cmidrule(lr){14-15}
    & NFE & Tok & NFE & Tok & NFE & Tok & NFE & Tok & NFE & Tok & NFE & Tok & NFE & Tok \\
    \midrule
    \rowcolor{gray!10} \multicolumn{15}{c}{\emph{Autoregressive Language Model}} \\
    \hspace{1.0mm}Qwen3-8B-Thinking & - & 3395 & - & 13950 & - & 14223 & - & 6448 & - & 13056 & - & 11533 & - & 10434 \\
    \midrule
    \rowcolor{gray!10} \multicolumn{15}{c}{\emph{Block Diffusion Language Model}} \\
    \hspace{1.0mm}SDAR-8B-Chat & 252 & 652 & 766 & 2303 & 454 & 1058 & 414 & 1088 & 142 & 328 & 467 & 924 & 416 & 1059 \\
    \hspace{1.0mm}DiRL-8B-Instruct & 1110 & 2322 & 3650 & 6867 & 3397 & 6187 & 2167 & 4313 & 382 & 764 & 1949 & 3427 & 2109 & 3980 \\
    \hspace{1.0mm}TraDo-8B-Instruct & 261 & 625 & 551 & 1190 & 513 & 1033 & 408 & 907 & 122 & 211 & 504 & 724 & 393 & 782 \\
    \hspace{1.0mm}TraDo-8B-Thinking & 3379 & 4787 & 11546 & 15216 & 11318 & 14951 & 6749 & 9191 & 10863 & 13244 & 8804 & 11429 & 8777 & 11470 \\
    \hspace{1.0mm}SDAR-30B-Thinking & 3557 & 5910 & 8646 & 14069 & 9440 & 15117 & 5324 & 8641 & 10100 & 13217 & 7624 & 10424 & 7449 & 11230 \\
    \midrule
    \hspace{1.0mm}TDAR-8B-Thinking (ours) & 1403 & 5590 & 4142 & 15589 & 4469 & 15648 & 2355 & 9424 & 5212 & 13937 & 2910 & 13510 & 3415 & 12283 \\
    \rowcolor{oursBlue!12}
    \hspace{2.5mm}+ BACD & 1317 & 6065 & 3842 & 15578 & 4123 & 15649 & 2175 & 9632 & 4758 & 14093 & 2701 & 13457 & 3153 & 12412 \\
    \rowcolor{RoyalBlue!30}
    \hspace{2.5mm}+ BACD + TCCF & 1490 & 5466 & 4458 & 17108 & 4768 & 16965 & 2483 & 9582 & 5490 & 15113 & 3419 & 14585 & 3684 & 13136 \\
    \midrule
    \hspace{1.0mm}TDAR-30B-Thinking (ours) & 1894 & 2933 & 7972 & 11983 & 8505 & 13086 & 3370 & 5268 & 9810 & 13091 & 6858 & 10035 & 6402 & 9399 \\
    \rowcolor{oursBlue!12}
    \hspace{2.5mm}+ BACD & 1805 & 2947 & 7469 & 11638 & 8326 & 13364 & 3201 & 5134 & 9371 & 12958 & 6597 & 9983 & 6128 & 9337 \\
    \rowcolor{RoyalBlue!30}
    \hspace{2.5mm}+ BACD + TCCF & 2002 & 3254 & 8193 & 12840 & 8810 & 14116 & 3357 & 5437 & 10174 & 14171 & 6943 & 10535 & 6580 & 10059 \\
    \bottomrule
\end{tabular}
    }
    \caption{Average number of function evaluations (NFE) and response token length (Tok) of different models and decoding methods on six reasoning benchmarks.}
    \label{tab:nfe_tokens}
\end{table*}

\section{Implementation Details}

\subsection{Model Training Configuration}
\label{appendix:implementation_training}

We transform Qwen3-8B-base into Block DLMs following the methodology of~\citet{cheng2025sdar}. Initially, we modify the generation mechanism through blockwise diffusion objective training ($B=4$) on a 50B-token annealing corpus subset, yielding TDAR-8B-Base ($B=4$).

Next, we conduct Supervised Fine-Tuning (SFT) on 3B long Chain-of-Thought (CoT) datasets employing our progressive block-size expansion approach. Starting from block size $4$, we gradually increase the training block size to 64 while maintaining consistent training configurations. Each stage undergoes 3 training epochs. Detailed training specifications are presented in Table~\ref{tab:training_config}.

\begin{table}[ht]
\centering
\begin{tabular}{lcc}
\toprule
\textbf{Config} & \textbf{CPT} & \textbf{SFT} \\
\midrule
Tokens        & 50B  & 3B   \\
Epochs        & 1    & 3    \\
Learning Rate & 1e-4 & 3e-5 \\
Batch Size    & 1024 & 128  \\
\bottomrule
\end{tabular}
\caption{Training configuration details for each stage.}
\label{tab:training_config}
\end{table}

\subsection{Inference Configuration}\label{appendix:implementation_inference}

We utilize SGLang\footnote{\url{https://github.com/sgl-project/sglang}} as our inference engine, which provides efficient support for both autoregressive and block diffusion decoding. All experiments are conducted on NVIDIA H200 GPUs.

We configure the maximum generation length to 30,000 tokens for SDAR-8B-Chat, DiRL-8B-Instruct, all TraDo-8B variants and our TDAR-8B-thinking model, which is sufficient for long Chain-of-Thought reasoning trajectories. We use a temperature of 1.0 to encourage diverse responses, while top-p and top-k sampling are not employed in our main experiments.

\section{Detailed Usage of TCCF Paradigm}
\label{appendix:detailed_usage_of_tccf}

The Think Coarse, Critic Fine (TCCF) paradigm implements a two-stage reasoning process that adaptively adjusts block sizes according to the functional roles of different reasoning segments. The paradigm exploits the observation that early-stage exploratory reasoning can tolerate coarse-grained generation, while later-stage refinement and verification require fine-grained precision. 
Algorithm~\ref{alg:tccf} presents the complete TCCF inference procedure.

\textbf{Stage 1: Coarse Thinking ($B_{think} = 16$):} 
In the first stage, the model performs exploratory reasoning with a large block size to efficiently generate the initial reasoning trajectory. The model processes the input prompt $p$ and generates reasoning content until it reaches the designated thinking boundary marker \texttt{</think>}. During this stage, the model rapidly explores the solution space, proposes potential approaches, and conducts preliminary analysis. The large block size ($B_{think} = 16$) enables efficient parallel decoding, significantly accelerating the exploration process.

\textbf{Transition Mechanism: } Once the model completes the thinking stage, the system automatically replaces this marker with a transition prompt: ``Let's check if there are any mistakes and give the final answer.'' This prompt serves as an explicit instruction to shift the model's focus from exploration to verification and refinement.

\textbf{Stage 2: Fine Critic ($B_{critic} = 4$): } 
In the second stage, the model switches to a smaller block size ($B_{critic} = 4$) to perform careful verification and refinement. The model reviews the reasoning generated in Stage 1, identifies potential errors, corrects mistakes, and produces the final answer with high reliability. The fine-grained decoding ensures precision in the critical refinement phase while maintaining computational efficiency overall, since Stage 2 typically generates significantly fewer tokens than Stage 1.

\begin{algorithm}[ht]
\caption{Think Coarse, Critic Fine (TCCF) Inference}
\label{alg:tccf}
\begin{algorithmic}[1]
\REQUIRE Input prompt $p$, model $\theta$, thinking block size $B_{think}$, critic block size $B_{critic}$, transition prompt $p_{trans}$
\ENSURE Final response $y$

\STATE \textcolor{blue}{// Stage 1: Coarse Thinking}
\STATE Initialize reasoning trajectory $r \leftarrow $ empty string
\STATE Set current block size $B \leftarrow B_{think}$
\WHILE{$r$ does not contain \texttt{</think>}}
    \STATE Generate next block: $r_{new} \sim p_\theta(\cdot | p, r; B)$
    \STATE Append to trajectory: $r \leftarrow r + r_{new}$
\ENDWHILE

\STATE \textcolor{blue}{// Transition: Replace marker with refinement prompt}
\STATE $r \leftarrow$ replace(\texttt{</think>} in $r$ with $p_{trans}$)

\STATE \textcolor{blue}{// Stage 2: Fine Critic}
\STATE Initialize final response $y \leftarrow r$
\STATE Set current block size $B \leftarrow B_{critic}$
\WHILE{not end-of-sequence}
    \STATE Generate next block: $y_{new} \sim p_\theta(\cdot | p, y; B)$
    \STATE Append to response: $y \leftarrow y + y_{new}$
    \IF{$y$ contains end-of-sequence token}
        \STATE \textbf{break}
    \ENDIF
\ENDWHILE

\STATE \textbf{return} $y$
\end{algorithmic}
\end{algorithm}

\begin{figure}[ht]
\begin{prompt}{\text{Transition prompt used in TCCF}}{case_study1}
Let's check if there are any mistakes and give the final answer.
\end{prompt}
\caption{Transition prompt used in TCCF paradigm to shift from coarse thinking ($B_{think} = 16$) to fine critic stage ($B_{critic} = 4$). This prompt is automatically inserted at the position where \texttt{</think>} marker appears in the Stage 1 output.}
\label{fig:case_study1}
\end{figure}

\textbf{Discussion on marker for boundary detection:} The TCCF paradigm is highly flexible and generalizable. Depending on the task requirements or model characteristics, the transition mechanism can be adapted using alternative prompt formulations (e.g., ``Review the steps above'' or ``Summarize the solution''), or by employing different structural markers. Our implementation provides a foundational framework for test-time scaling in Block DLMs, offering a broad design space for future exploration in efficient and reliable long-context reasoning.

\section{Case Study}
\label{appendix:case_study}

Figure~\ref{fig:case_study2} presents a detailed case study demonstrating why the TCCF paradigm is crucial for complex reasoning tasks. The problem asks for the least possible sum of distinct positive integers with product 84, which requires searching through multiple factor combinations.

\textbf{Stage 1: Efficient Exploration via Coarse Thinking ($B=16$).} 
In this phase, the model acts as a rapid explorer. By using a large block size, it can quickly traverse the reasoning path and identify high-probability solutions. As shown in the case, the model correctly identifies the factorization of 84 and proposes several valid sets, arriving at a suboptimal answer of 15. Although the coarse granularity ($B=16$) accelerates the generation of the reasoning chain, it inherently carries a trade-off in precision, leading the model to overlook the optimal combination $\{3, 4, 7\}$ in the vast search space. This stage provides the necessary \textit{breadth} of reasoning without consuming excessive computational time.

\textbf{Stage 2: Precision Correction via Fine Critic ($B=4$).} 
The transition prompt (``Let's check...'') triggers a mode switch, acting as a verifier. By switching to the finest granularity ($B=4$), the model allocates its full computational capacity to scrutinize the previous conclusion. In this high-precision mode, the model successfully detects the oversight from the coarse stage and discovers the subtle combination $\{3, 4, 7\}$ (sum=14). 

\textbf{Effectiveness Analysis:} 
This case illustrates the core advantage of TCCF. It decouples reasoning generation from answer verification. Pure coarse decoding ($B=16$) is fast but prone to detail errors (finding 15 instead of 14), while pure fine decoding ($B=4$) is accurate but computationally expensive. TCCF using the first stage to efficiently narrow down the solution space and the second stage to ensure the final answer is precise. This coarse-to-fine mechanism allows the model to correct its own hallucinations or oversights, directly contributing to the accuracy improvements observed in our benchmarks.

\begin{figure*}[ht]
\begin{prompt}{\text{Model Output}}{case_study2}
\small

\textbf{Problem:} The product of a set of distinct positive integers is 84. What is the least possible sum of these integers?

\textbf{Stage1 Output:}

<think>

We need to find a set of distinct positive integers (i.e., the integers are all different) whose product is 84, and we want to minimize the sum of the integers.

First, note that factorization of 84 = 2\textasciicircum2 * 3 * 7...

Potential sets:

- Single integer: {84} sum = 84.

- Two numbers: distinct positive integers a, b, a*b=84... minimal sum among two numbers without 1 is 19 (7+12).

- Three numbers: a*b*c=84, distinct >1... We found such: {2,3,14} sum=19; {2,4,21} sum=27; {2,6,14} sum=22; {2,7,12} sum=21; {3,4,14} sum=21. So minimal sum among three numbers is 19.

- Four numbers: ... we could have {4,3,7,1} product = 4*3*7*1 = 84... Sum = 4+3+7+1 = 15.

Thus the minimal sum among sets of size 4 is \boxed{15}. 

Now we need to produce final answer.

\textcolor{blue}{\textbf{Let's check if there are any mistakes and give the final answer.}}

\textbf{Stage2 Output:}

Let me pause and think about this more carefully... But we need to also consider if there is any other set with sum 14? Let's see if there's any set of size >1 with sum 14? The set {1,3,4,7} sum = 15. No, sum 14 would require {1,3,4,6}? Product = 72 not 84...

But I should also consider whether the problem intends the set to contain at least two integers... However, the set with sum 14 includes 3 numbers, all distinct >0, product 84. So if we allow any cardinality >= 2, the minimum sum is 14. But we need to verify if there is a set of size 2 with sum less than 14. The minimum sum among pairs not including 1 is 19. So no. So if cardinality >=2, the answer is 14. But we have a set of size 3 and sum 14. So answer would be 14, not 15.

But we need to see what the problem expects... Actually 14 < 15. But we found a set of three numbers with sum 14 (3+4+7). So sum 14 is less. So the minimal sum must be <=14. But we found a set {3,4,7} sum = 14, product = 84. So sum 14 is possible.

Wait we need to double-check: Are there any sets of size >=4 with sum <14? Possibly {1,3,4,7} sum = 15 which is higher than 14. So minimal sum is 14...

But earlier we thought of including 1 to reduce sum further. The minimal sum is 14 from {3,4,7}. So answer is 14.

Thus final answer: \boxed{14}. Let's produce final answer accordingly.

</think>

**Final Answer:**

The prime factorisation of 84 = 2²·3·7.  
Considering all possible sets of distinct positive integers with product 84:

- One integer: {84} → sum = 84

- Two integers: minimal sum = 7+12 = 19

- Three integers: minimal sum = 3+4+7 = 14 (set {3,4,7})

- Four integers: must include 1, minimal sum = 1+3+4+7 = 15

- Five or more integers: impossible  

The smallest achievable sum is \boxed{14}.

\end{prompt}
\caption{Case study of a representative problem from the Math500 benchmark.}
\label{fig:case_study2}
\end{figure*}

\section{Detailed Decoding Configurations and Algorithms}
\label{appendix:decoding_details}

In this section, we provide formal formulations and hyperparameter configurations for the mainstream DLM decoding algorithms evaluated in Section~\ref{sec:analysis_decoding_algorithm} and Table~\ref{tab:decoding_comparison}.

\subsection{Static Decoding}
Static decoding methods utilize a fixed step-budget schedule that is independent of token prediction confidence.
\begin{itemize}
    \item \textbf{StaticConfidence:} In each denoising step within a block of size $B$, a fixed number of tokens $n = \lceil B / N \rceil$ are unmasked, where $N$ is the total pre-allocated denoising steps. The tokens with the highest confidence (Equation~5) are selected. Under our evaluation setup, we set $N = 4$ for a block size $B = 16$.
    \item \textbf{StaticEntropy:} Similar to StaticConfidence, this method unmasks a fixed $n$ tokens per step but uses the negative entropy of the predicted token distribution as the selection metric. The entropy at position $i$ is defined as $H_i = -\sum_{v \in \mathcal{V}} p_i(v) \log p_i(v)$, and the $n$ positions with the lowest entropy are unmasked at each step.
\end{itemize}

\subsection{Dynamic Confidence Decoding}
Dynamic Confidence Decoding~\citep{wu2025fastdllm} serves as our default baseline and adapts the number of generated tokens dynamically. At each step $t$, the candidate set of tokens to be unmasked is defined as:
\begin{equation}
    \mathcal{D}_t = \{i \in \mathcal{M}_t^k \mid c_i(\mathbf{x}_t^k) > \tau\}
\end{equation}
where $\tau$ is a constant threshold. If $\mathcal{D}_t = \emptyset$, the algorithm falls back to unmasking the single token with the highest confidence score to guarantee convergence. In our experiment, we set the default threshold $\tau = 0.9$ following official implementations.

\subsection{EB-Sampler}
Introduced by~\citet{benhamu2025acceleratedsamplingmaskeddiffusion}, the \textbf{EB-Sampler} is an adaptive multi-token unmasking strategy designed for masked diffusion models. Its central insight is that a partially unmasked sequence often deterministically determines multiple masked tokens, allowing aggressive parallel unmasking without sacrificing generation quality.

At each step, EB-Sampler dynamically estimates the cumulative error of predicting multiple tokens and unmasks them simultaneously, provided that the total error proxy remains below a predefined budget $\gamma$. We evaluate three distinct error proxy variants of EB-Sampler:
\begin{itemize}
    \item \textbf{Confidence Proxy:} Measures error using the predicted probability of the argmax token, ensuring that the joint confidence approximation remains bounded.
    \item \textbf{Entropy Proxy:} Uses the predicted entropy at each position to gauge the information gain and uncertainty.
    \item \textbf{Margin Proxy:} Gauges prediction certainty by calculating the difference between the top-1 and top-2 predicted token probabilities: $\text{Margin}_i = p_i(v^{(1)}) - p_i(v^{(2)})$.
\end{itemize}
We configure the error budget to $\gamma = 0.5$ for all three variants in our benchmarks.

\subsection{Saber}
The Saber algorithm, proposed by~\citet{dong2025saber}, addresses the brittleness of DLMs in structured generation tasks (e.g., code reasoning) by introducing a training-free backtracking-enhanced remasking solver. Saber implements two core mechanisms:
\begin{enumerate}
    \item \textbf{Adaptive Acceleration:} The solver dynamically adjusts the number of unmasked tokens at each step according to the model's evolving confidence distribution.
    \item \textbf{Backtracking Enhanced Remasking:} As generation proceeds and new context is established, Saber re-evaluates the confidence of previously unmasked tokens. If a token's confidence drops below a dynamic backtracking threshold, Saber applies a re-masking operation (reverting the token back to \texttt{[MASK]}), thereby pruning compounding errors in early generation steps.
\end{enumerate}
For our evaluation on BDLMs, we configure the momentum-based early termination coefficient $\mu = 2$.

\section{Block Size Selection via Knee Point Analysis}
\label{app:block_size_knee_point}

\begin{figure*}[t]
    \centering
    \includegraphics[width=\linewidth]{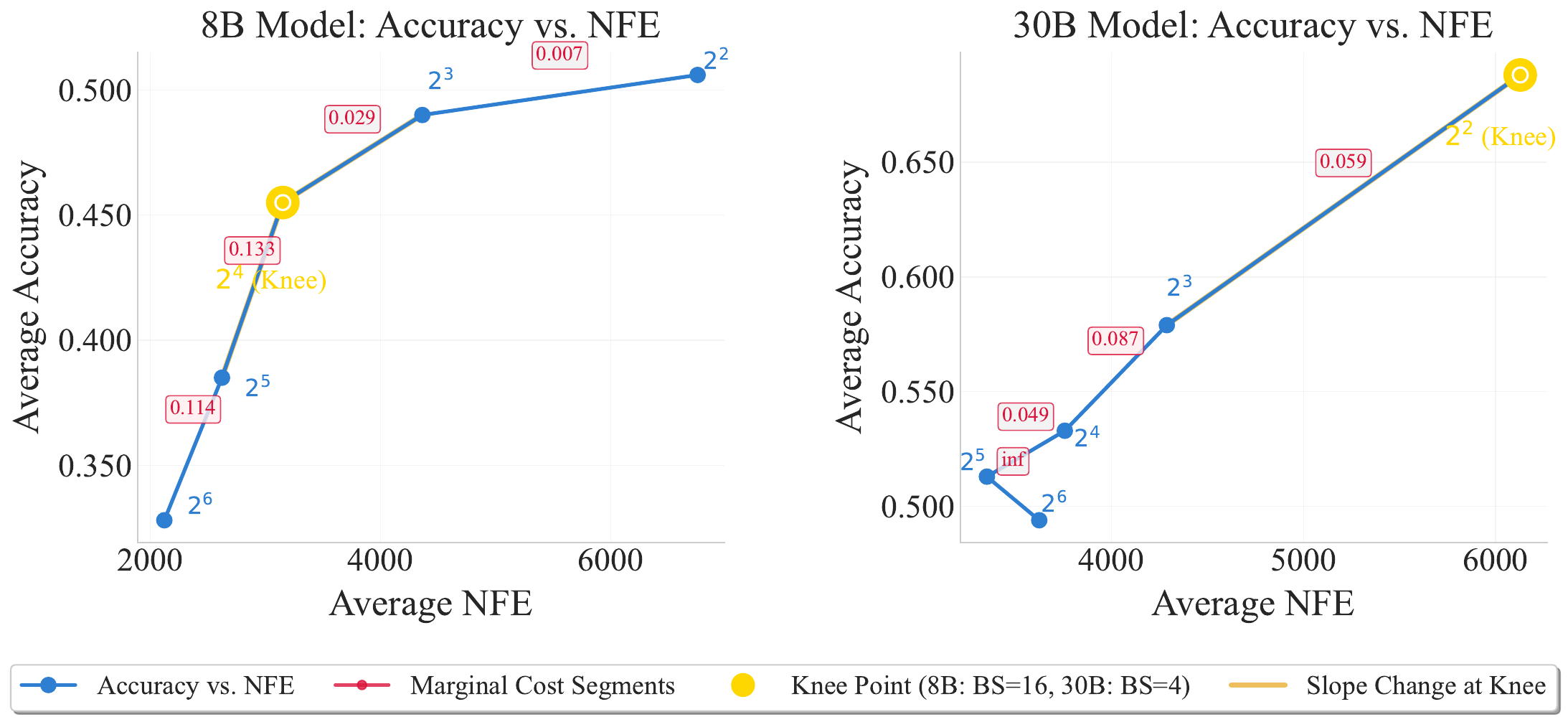}
    \caption{Accuracy vs.\ NFE trade-off for different block sizes. Numbers on each segment denote the marginal cost (accuracy gain per 1{,}000 additional NFE). The golden circle marks the knee point where the marginal return drops sharply.}
    \label{fig:block_size_knee_point}
\end{figure*}

We determine the optimal block size by identifying the \emph{knee point} of the Accuracy--NFE curve, i.e., the point beyond which further NFE investment yields diminishing accuracy gains.

\paragraph{8B Model.}
As shown in Figure~\ref{fig:block_size_knee_point} (left), the marginal cost drops by 4.6$\times$ from the segment $B{=}32{\to}16$ (0.133) to $B{=}16{\to}8$ (0.029), indicating that $B{=}16$ is the knee point where additional NFE provides substantially less accuracy improvement.

\paragraph{30B Model.}
For the 30B model (Figure~\ref{fig:block_size_knee_point}, right), the marginal cost peaks at $B{=}16{\to}8$ (0.087) and declines to 0.059 at $B{=}8{\to}4$, identifying $B{=}4$ as the knee point. We therefore adopt $B{=}16$ for the 8B model and $B{=}4$ for the 30B model throughout our experiments.